\documentclass[acmsmall,screen]{acmart}
\renewcommand\footnotetextcopyrightpermission[1]{} 
\AtBeginDocument{%
  \providecommand\BibTeX{{%
    \normalfont B\kern-0.5em{\scshape i\kern-0.25em b}\kern-0.8em\TeX}}}

\setcopyright{acmcopyright}
\copyrightyear{2018}
\acmYear{2018}
\acmDOI{10.1145/1122445.1122456}

\acmJournal{JACM}
\acmVolume{37}
\acmNumber{4}
\acmArticle{111}
\acmMonth{8}

\usepackage{indentfirst}
\usepackage{color}
\usepackage{url}

\begin{document}

\title{Aesthetic Visual Question Answering of Photographs}

\author{Xin Jin}
\email{jinxinbesti@foxmail.com}
\affiliation{%
	\institution{Beijing Electronic Science and Technology Institute}
	\city{Beijing}
	\state{Beijing}
	\country{China}
}

\author{Wu Zhou}
\email{zhouwu\_nj@126.com}
\affiliation{%
	\institution{Beijing Electronic Science and Technology Institute}
	\city{Beijing}
	\state{Beijing}
	\country{China}
}

\author{Xinghui Zhou$^{*}$}
\email{graydove@mail.ustc.edu.cn}

\affiliation{%
	\institution{University of Science and Technology of China}
	\city{Hefei}
	\state{Hefei}
	\country{China}
}

\author{Shuai Cui}
\email{shucui@ucdavis.edu}
\affiliation{%
	\institution{University of California, Davis}
	\city{Davis}
	\country{USA}
}

\author{Le Zhang}
\email{lezhang.thu@gmail.com}
\affiliation{%
	\institution{Beijing Electronic Science and Technology Institute}
	\city{Beijing}
	\state{Beijing}
	\country{China}
}

\author{Jianwen Lv}
\email{513415184@qq.com}
\thanks{*Corresponding authors}
\affiliation{%
	\institution{Beijing Electronic Science and Technology Institute}
	\city{Beijing}
	\state{Beijing}
	\country{China}
}

\author{Shu Zhao}
\email{zhaoshu0104@163.com}
\affiliation{%
	\institution{Beijing Electronic Science and Technology Institute}
	\city{Beijing}
	\country{China}
}



\begin{abstract}
   Aesthetic assessment of images can be categorized into two main forms: numerical assessment and language assessment. Aesthetics caption of photographs is the only task of aesthetic language assessment that has been addressed. In this paper, we propose a new task of aesthetic language assessment: aesthetic visual question and answering (AVQA) of images. If we give a question of images aesthetics, model can predict the answer. We use images from \textit{www.flickr.com}. The objective QA pairs are generated by the proposed aesthetic attributes analysis algorithms. Moreover, we introduce subjective QA pairs that are converted from aesthetic numerical labels and sentiment analysis from large-scale pre-train models. We build the first aesthetic visual question answering dataset, AesVQA, that contains 72,168 high-quality images and 324,756 pairs of aesthetic questions. Two methods for adjusting the data distribution have been proposed and proved to improve the accuracy of existing models. This is the first work that both addresses the task of aesthetic VQA and introduces subjectiveness into VQA tasks. The experimental results reveal that our methods outperform other VQA models on this new task.
\end{abstract}
\keywords{ Multimodal learning, Aesthetic computing, Visual question and answering, Semi-automatic labeling, Transformer}

\keywords{aesthetic mixed dataset with attributes, multitasking, external attribute features, ECA channel attention}

\maketitle

\section{Introduction}
In the researches of computer vision and natural language processing, researchers always focus on direct and obvious goals instead of image aesthetics with different attributes. In the past decade, many researchers were interested in image aesthetics. However, the popular methods were to score images and describe attributes of images. Either giving a score ranging from 0 to 10 to evaluate aesthetics of images or describing a images' attributes\cite{jin2019aesthetic}, such as color, lighting, and composition, in one sentence are limited and lack description in details. 

The state-of-the-art model of visual question answering (VQA) has achieved multiple purposes on many large-scale datasets. According to the category of the datasets, the subtask of VQA can be divided into task-based, inference-based, and text-based, like TextVQA \cite{singh2019towards} dataset. Many famous models \cite{li2020oscar,chen2020uniter,li2019visualbert} have been developed in these visual question answering tasks. However, they lack the overall investigation and reasoning of the picture.

Many attribute labels related to image aesthetics can be collected in photography (for example, image classification and evaluation labels available in subjectively scored datasets). Although some studies have used image tags for regression, multi-task evaluation, and image captioning, they ignore the available information of images in multiple information dimensions.

\begin{figure}[htb]
	\centering
	\includegraphics[width=0.7\textwidth]{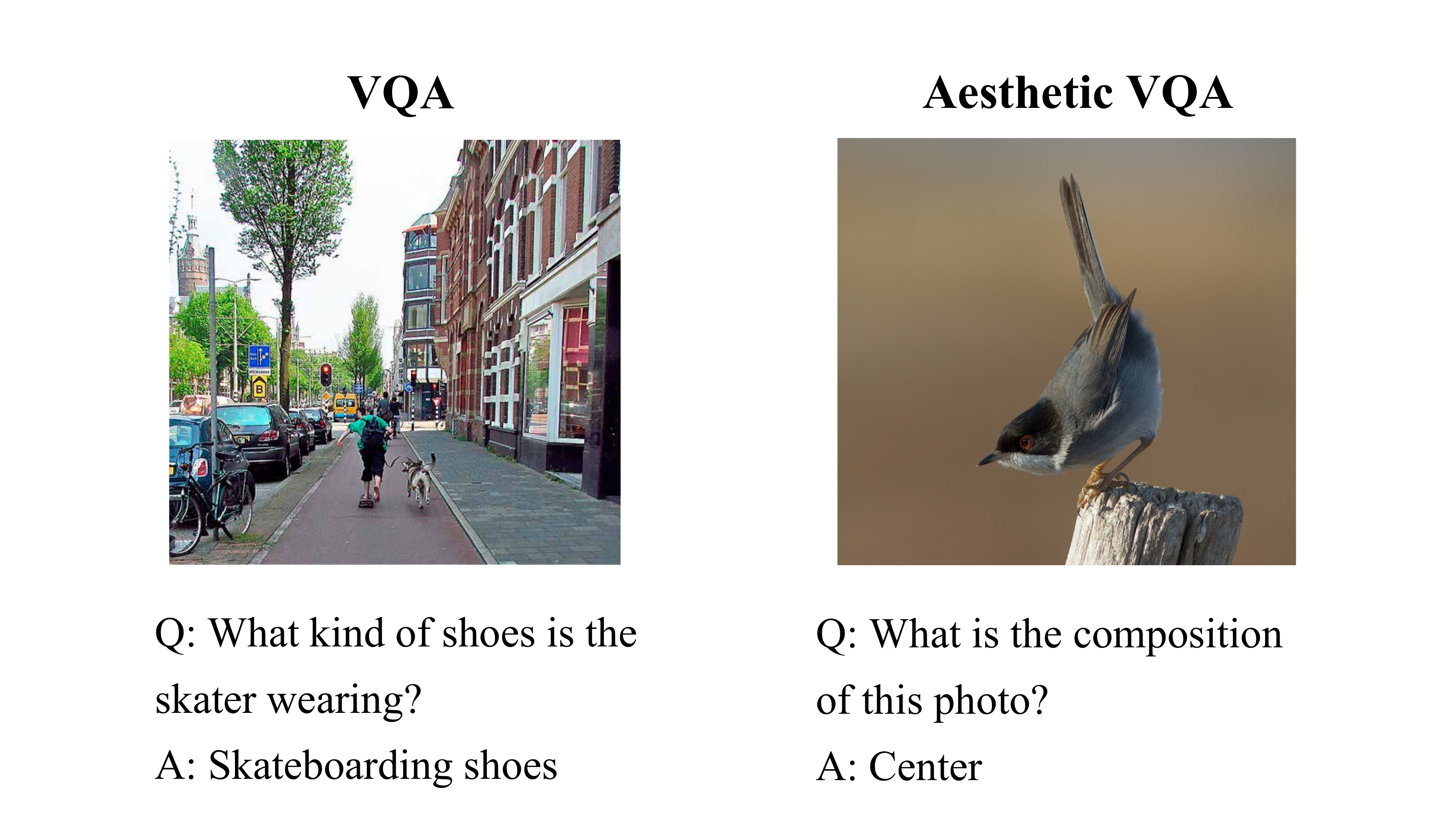}
	\caption{VQA and Aesthetic VQA. The AVQA focuses on the overall aesthetic attributes of the image such as shot, lighting and colors.}
	\label{fig:AVQA}
\end{figure}

Some large-scale visual question answering datasets, such as VQA2.0 \cite{balanced_vqa_v2} and Visual Genome \cite{krishna2017visual}, include some questions related to the overall image; GQA \cite{hudson2019gqa} and NLVR2 \cite{suhr2017corpus} include some questions about designing image content reasoning. By a lot of computational cost and testing, we can get the answer related to the question about overall characteristics of the image. There is no dataset that focuses on visual question answering based on the overall image.

Most of current VQA researches focus on objective QA pairs. This paper mainly addresses aesthetic visual question answering (AVQA) of images, as shown in Figure \ref{fig:AVQA}. This task is both aesthetic and subjective; also, it is a complementary task for the previous QA studies. This problem is important since subjective questions and answers are very prevalent among human discourses. In this paper, we present a dataset with aesthetic QA pairs, aesthetic visual question answering (AesVQA). Images in the new dataset are all marked with aesthetic labels through series of multiple computer vision sub-tasks. These labels cover composition, color, subject, lighting, genres, techniques and emotions of photos. 

To obtain reliable labels, we develop some unsupervised and semi-supervised computer vision methods based on image processing and large restraint model. At the same time, we propose a threshold function and design a small amount of manual labelling image adjustment function to get a uniform and credible dataset distribution. 

In particularly, the labels of lights and colors are designed based on the specific objects in the image. The labels of scene and composition are designed based on the whole picture. The other labels based on subjective evaluations of images, especially techniques and emotions, provide extra dimensional information in question and answering. The task becomes more challenging by adding some human psychological evaluations. The main contributions of this paper can be summarized as: 

\begin{itemize}
	\item {\verb|New Task|}: The new task of image aesthetics question and answer, and it is the first work that introduces subjective answering to VQA problems.
	\item {\verb|New Labels|}: Semi-automatic QA labelling of images gets more information when answering and methods of dataset distribution adjustment.
	\item {\verb|New Points of VQA|}: The traditional visual question answering method based on objective detection is optimized to half-object-detection-based (such as lighting and color labels) and half-whole-image-based (such as subject and composition labels).
\end{itemize}

\section{Related work}

Aesthetics and computer vision are inextricably linked, and people are demanding higher quality of image data. How to process the images into a form that is more compatible with human preferences has become a critical issue. It is feasible to use computer technology to study the aesthetics of images\cite{lu2015rating}.

The earliest image aesthetic quality evaluation was only the two-category evaluation of scores and good or bad, based on which a considerable number of scholars have made considerable contributions on how to score and score on multiple attributes of images. The earliest study of aesthetic image quality evaluation is this paper\cite{chang2017aesthetic}.

Visual question answering is a difficult problem that straddles computer vision and natural language processing, and its task requires the extraction of not only image features but also textual partial features. Recent VQA datasets such as DAQUAR\cite{ren2015exploring}, Visual7W\cite{zhu2016visual7w}, Visual Madlibs\cite{yu2015visual}, FM-IQA\cite{gao2015you}, VQA\cite{balanced_vqa_v2}, etc. are available. Unlike look-and-talk tasks, simply fusing image and text features often does not yield the desired features, i.e., answers.

Visual question answering requires the model to process from image and language to combine and give the answer effectively. So, VQA is both interdisciplinary and challenging. Also, due to characteristics of image aesthetics, there is no professional image aesthetics Q \& A(Question and Answer) dataset. Therefore, it is important to propose a common aesthetic Q \& A datasets. By the visual model, we use classical Q\& A datasets as reference to generate our dataset. The classic Q \& A datasets are always generated by machines, such as: VQA\cite{balanced_vqa_v2}, TextVQA\cite{singh2019towards}, EST-VQA\cite{wang2020general} and VizWiz-VQA\cite{bigham2010vizwiz}.

One of the difficulties of the image aesthetics task is the disadvantage of inaccurate subjective evaluations but low number of objective ones. Borrowing a large amount of data as a foundation, it will be possible to mine a sufficient number of images and corresponding comments with a high enough standard, and then further convert the comments into the desired Q\&A pairs. Proxy objectification with statistical features of subjective evaluation is a common approach in current image aesthetics tasks.

To date, there is no comprehensive aesthetic VQA dataset. Previous research work proposed the AQUA\cite{garcia2020dataset} dataset, which focuses more on studying the artistic aspects of paintings, while our dataset focuses more on the aesthetic aspects of photographs. In previous studies, researchers would apply attention mechanisms to images or text to obtain better results, but it is difficult to obtain the desired features from the many image features because of the different feature spaces and the presence of tensor features with ultra-high dimensionality of images.
   
\begin{table*}[t]
	\centering
	\caption{VQA datasets' answer words length accounted for the proportion of all answers table. Answers with two words or more are 94.6\% in AesVQA dataset, which leads a more difficult VQA task.}
	\begin{tabular}{c|ccc|c}
		\hline
		& \multicolumn{3}{c|}{The Length of The Answer Words} &                        \\ \hline
		Dataset       & One Word  & Yes/No & Two Words or More  & The Number of QA Pairs \\ \hline
		VQAv2 \cite{balanced_vqa_v2}         & 51.2\%    & 45.0\%             & 3.8\%                & 1,105,904                \\ \hline
		Visual7W \cite{zhu2016visual7w}      & 51.2\%    & 21.2\%		            & 27.6\%               & 139,868                 \\ \hline
		Visual Genome \cite{krishna2017visual} & 52.5\%    & 21.2\%		           & 26.3\%               & 1,445,233                \\ \hline
		AesVQA  & 5.4\%     & \textbf{0\%}		 & \textbf{94.6\%}                 & 324,756 \\ \hline
	\end{tabular}
	\label{tab:AesVQA}
\end{table*}

\begin{figure*}[htbp]
	\centering
	\includegraphics[width=0.8\textwidth]{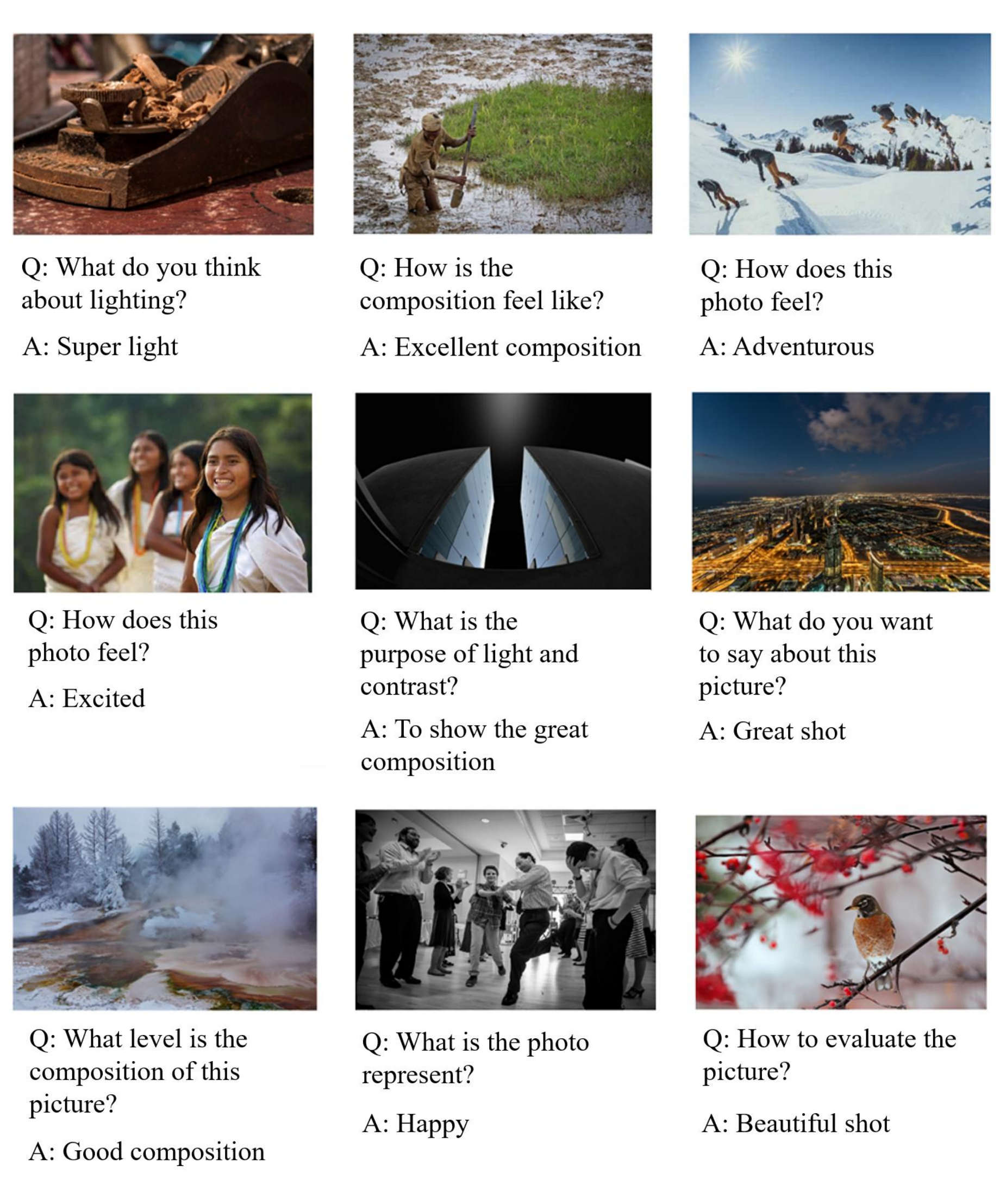}
	\caption{Pictures, Questions and Answers in the Aesthetic VQA dataset.}
	\label{fig:aesqadataset}
	\label{fig:dataset}
\end{figure*}

\section{Aesthetic Visual Question Answering}
To study the answers to questions related to image aesthetics and the overall content of the image, and to reduce the cost of manual annotation, we proposed AesVQA, as shown in Figure \ref{fig:aesqadataset} and Table 1. This dataset is an image aesthetic question and answer dataset based on an unsupervised model. We will start by describing how to select the images used in AesVQA. Then, we explain our data collection pipeline to collect questions and answers.

\begin{figure*}[t]
	\centering
	\includegraphics[width=1.0\textwidth]{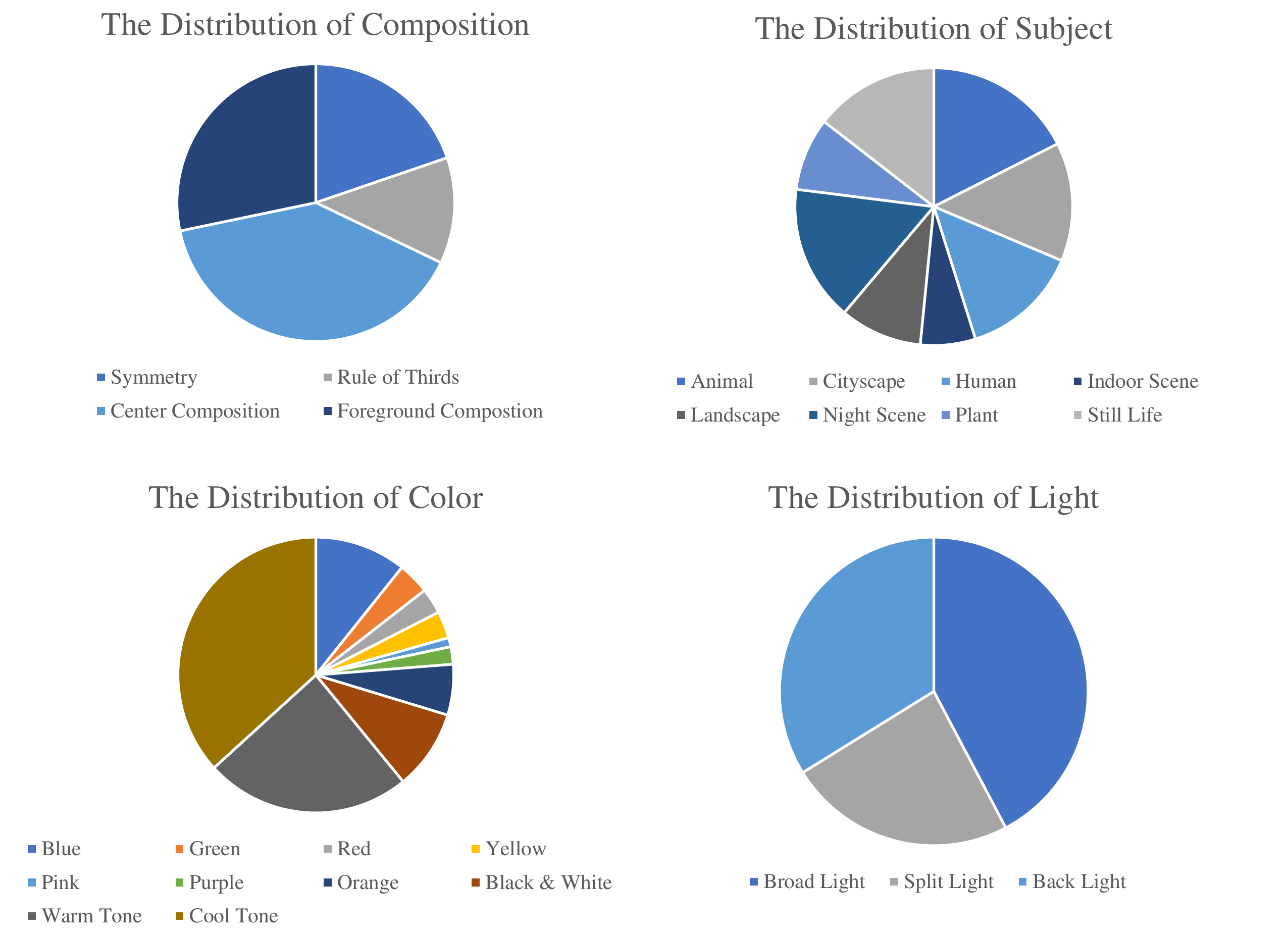}
	\caption{The distribution of the basic aesthetic labels in the AesVQA dataset.}
	\label{fig:distribution}
\end{figure*}
\begin{figure*}[t]
	\centering
	\includegraphics[width=1.0\linewidth]{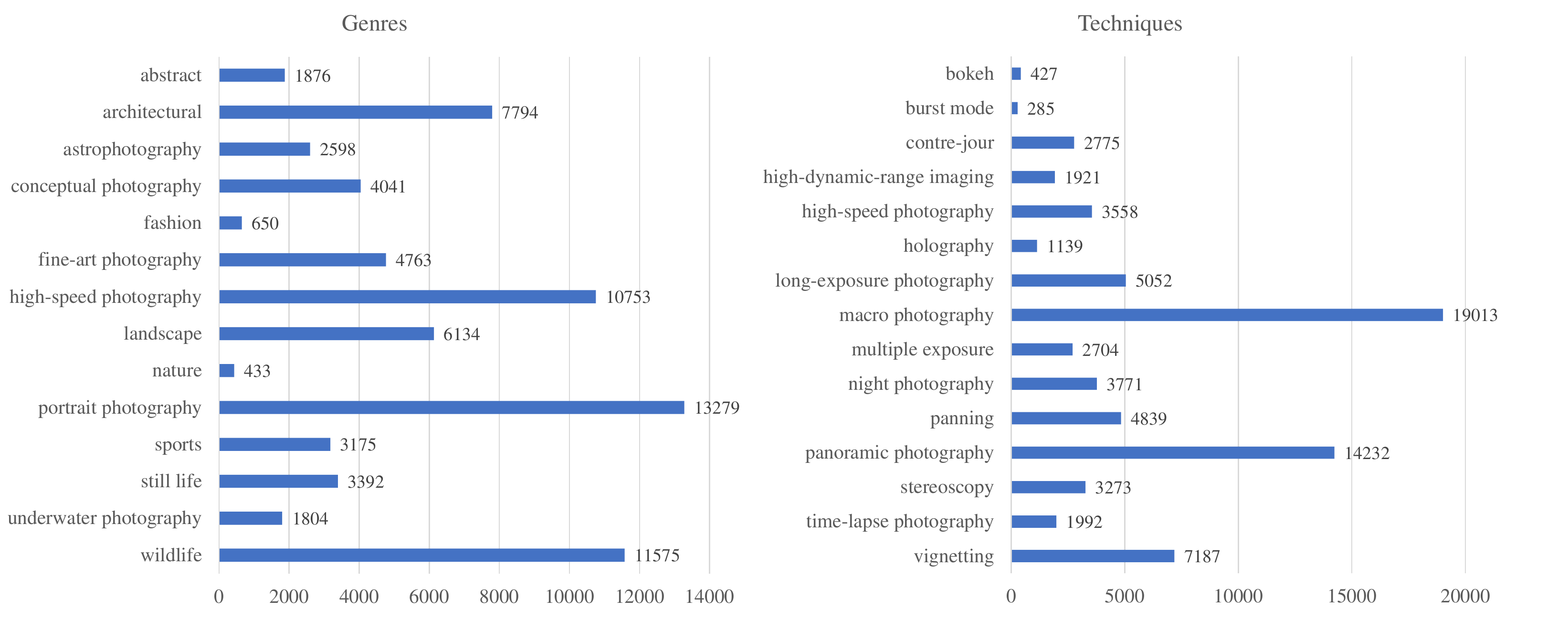}
	\caption{The distribution of the genres labels in the AesVQA dataset.}
	\label{fig:genres}
\end{figure*}

\begin{figure}[t]
	\centering
	\includegraphics[width=\linewidth]{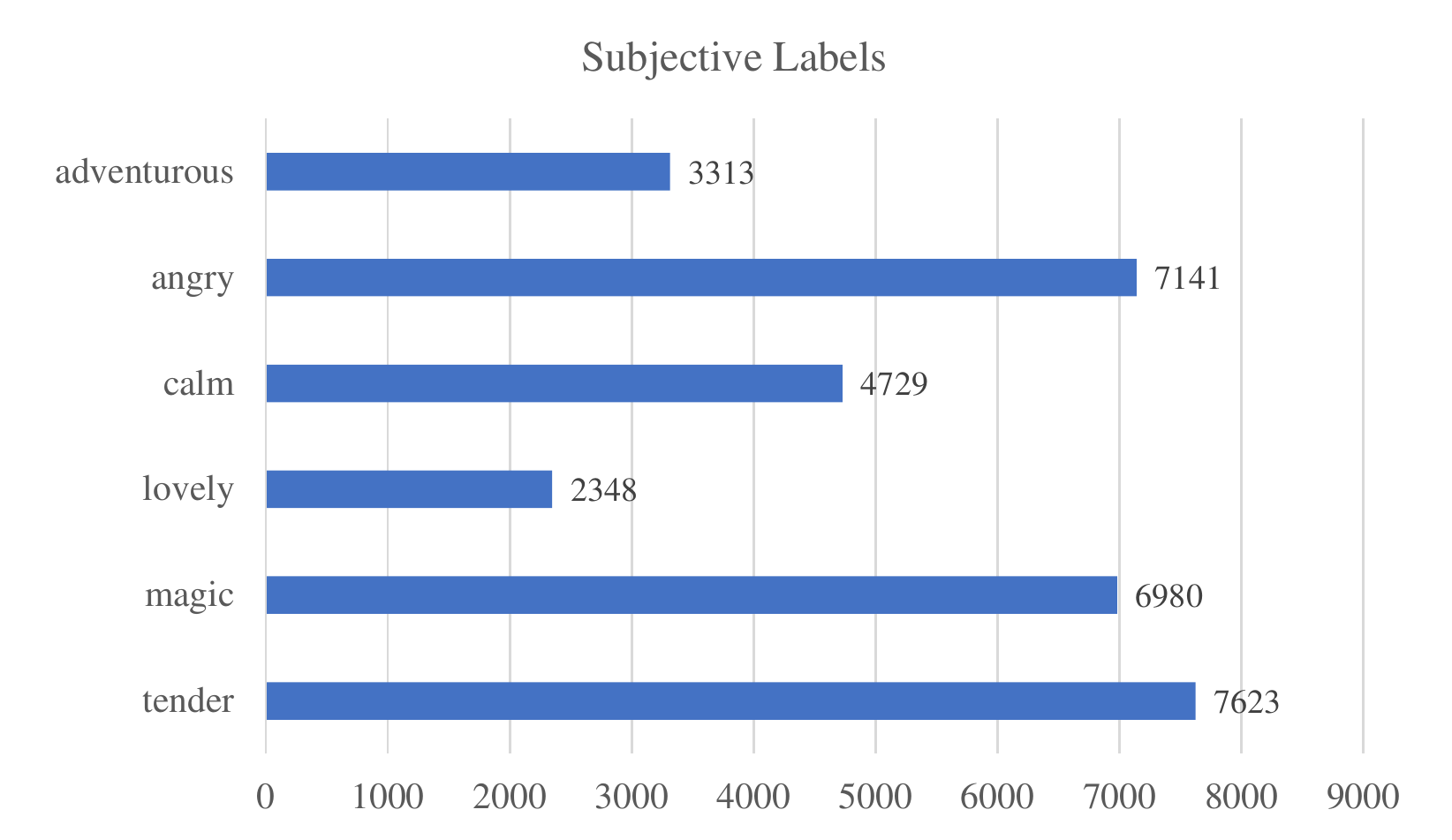}
	\caption{The distribution of photos's subjective labels in the AesVQA dataset. Including tender, magic, lovely, calm, angry, adventurous and many other attributes.}
	\label{fig:subjective}
\end{figure}

\subsection{Images of AesVQA}
We use pictures from the Flickr website as our source pictures, which conforms to developing a VQA model based on image aesthetics and the overall content of the image. We are most interested in pictures in categories such as landscapes, people, still lifes, and animals. Several categories in the Explore section of the Flickr website meet this condition.

To obtain accurate images, to determine the aesthetic information of the image and the overall characteristics of the image, we deleted the pictures with borders and the pictures with a too large aspect ratio (if the aspect ratio of the picture is greater than 1:2, delete it). We use the OCR model Rosetta on these images to delete some images that contain watermarks that may be involved in the infringement.

To automate this process to identify categories that tend to contain images with text, we select 100 random images for each category (if the maximum number of images for that category is less than 100, select all images). We run the state-of-the-art OCR model Rosetta [6] on these images and calculate the average number of OCR boxes in the category.

Based on the 190,000 Flickr images we crawled, after artificially filtering out abstract images and unrealistic images from image photography, filtering these (and a small amount of noisy data from manual annotations) can obtain 72168 images. image. Manual screening requires that the image must be taken from the real world, not obtained through PS or other software editing tools.

\subsection{Generation and filtering of AesVQA}
In order to generate question and answer pairs, we designed our own model. Our model based on sequence to sequence method, including three modules: text content encoder, answer encoder and decoder. In addition, the quality of the generated AesVQA data is uneven, and there is a big gap with the common VQA dataset. Therefore, the data needs to be screened again after QA is generated. We propose an aesthetic question and answer filtering method using LDA\cite{blei2003latent}.

The text content coder is used to process the input comments. At the same time, it needs to calculate the attention weight of different words and the weight of specific words in the replication mechanism. The answer encoder is used to process the selected answer words or phrases, filter the words according to the attention weight, and encode the words into a format suitable for the decoder. The decoder is responsible for using the cyclic neural network to process the input tensor data, and generating questions according to the weight data between the answer and the comment. 

For the generated question and answer pairs, the manual filtering method is used to remove the inappropriate question and answer pairs, so as to ensure that at least one question and answer pair generated by each picture is reasonable. These question and answer pairs are formed into a small text set, and the correlation between the text set and multiple topic words obtained from previous LDA topic calculation is calculated, which is used as the threshold to evaluate whether a question and answer pair meets the aesthetic question and answer model.

Calculate the relevance between the remaining sentences and the topics selected by LDA, and judge whether the relevance of the selected sentences is greater than the topic threshold. When the relevance value of the selected sentence is greater than the threshold, it is determined as a question and answer pair that meets the requirements, otherwise it is determined as a question and answer pair that does not meet the requirements.

Finally, since the quality of the generated issues is not very high, the solution is to make manual corrections.

\subsection{Basic Aesthetic Labels}

The first part of the AADB\cite{kong2016photo} dataset we constructed is the basic image aesthetic tags, including composition, color, subject, and light. Among them, in order not to reduce the VQA task to a simple classification task, and to ensure that the specific image attributes are associated with specific objects, we design light labels and color labels to be related to the content of the image, while composition labels and subjects labels are related to the overall image. Here, the whole image is defined as multiple targets or content obtained by the fusion of background and foreground information. The biggest feature of this content is that each part of the image is required to obtain it, rather than part-based objects.

Among them, the symmetry of the composition labels is obtained by the SSIM\cite{wang2004image} calculation of the image. When the left and right parts of the image or the upper and lower parts of the image after the mirror symmetry, the SSIM is higher than the threshold, and it is judged to be an asymmetric image. The rule of thirds and the center composition use the faster R-CNN based on the visual genome to obtain the position of the most distinctive object in the image. Midas is used for the foreground composition, and the image is divided into three parts: the upper, middle, and lower parts, and judge whether there is a significant difference in the depth of field between the upper part and the lower part.

Color labels depend on the type and color of objects in the image obtained by faster R-CNN. To enrich the diversity of colors, we have added warm tone and cold tone, using target detection to cut out the selected objects in the image, and classify their average colors, and judge based on the correlation between the average color and other known main colors warm tone or cold tone.
The attributes of the subject are labeled using the CLIP\cite{radford2021learning} pre-training model. According to the classification results of the subject on the AVA\cite{murray2012ava}, it can be known that CLIP can deal with the category labeling of multiple subjects. At the same time, to make up for the errors that may be caused by automatic marking, some manually marked image attributes are subsequently used to correct the results.

The attribute of Light is judged by judging the grayscale histogram of the image. When the image is not black and white, but still has more black parts, it is marked as backlight; if the histogram part of the image is on both sides, it is marked as split light; If the histogram of the image is evenly distributed, it is marked as broad light. Similar to the Subject attribute, this part also uses some artificially marked image attributes to correct the result.

\begin{figure*}[t]
	\centering
	\includegraphics[width=0.9\textwidth]{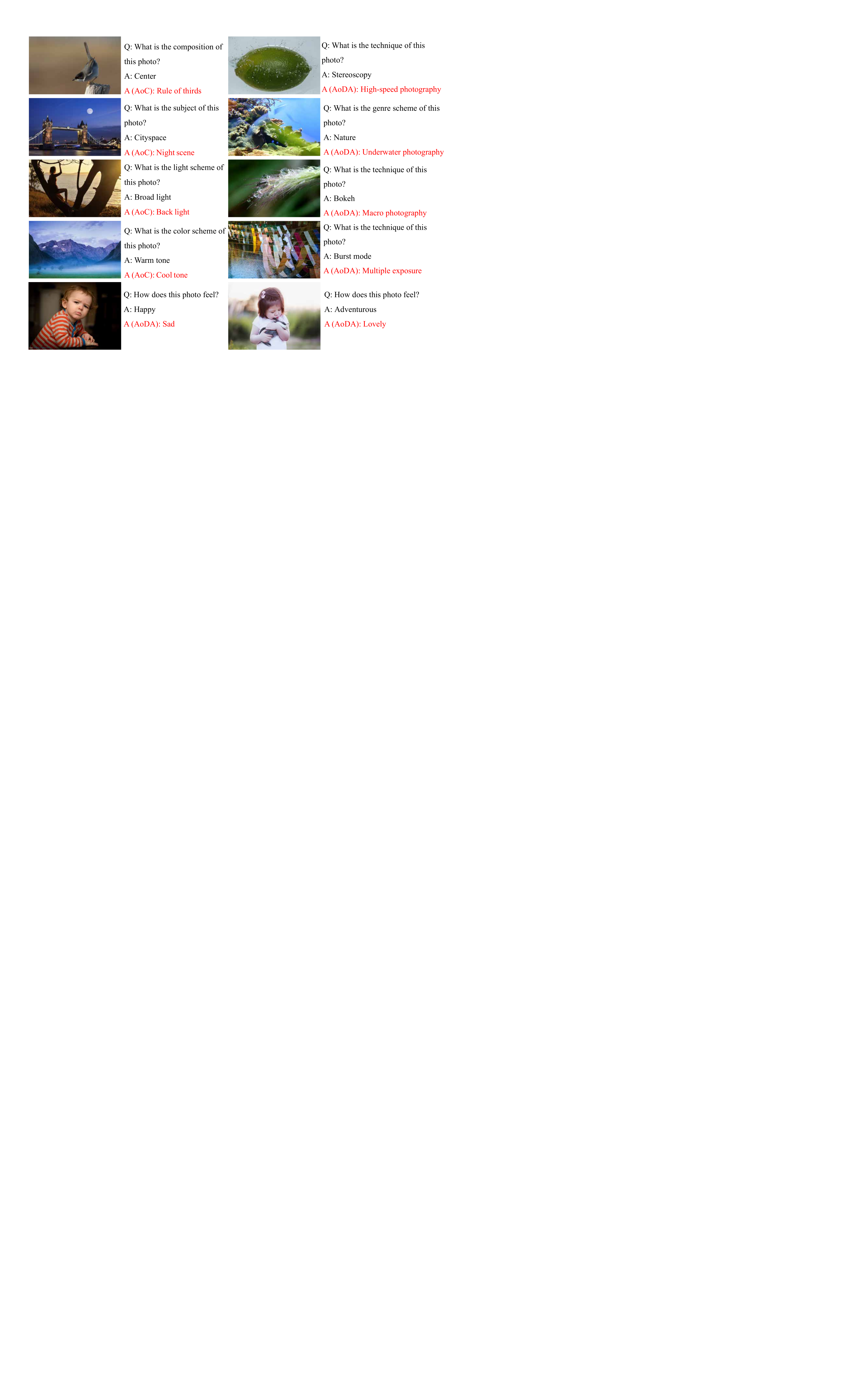}
	\caption{Question answers which before and after the adjustment of image's confidence and the adjustment of the distribution of answers.The \textbf{AoC} means the adjustment of image's classify confidence. The \textbf{AoDA} means the adjustment of distribution of answers. These two tricks will help the model achieving a higher accuracy.}
	\label{fig:photo}
\end{figure*}

\subsection{Distribution in VQA Dataset}

The distribution of the dataset is extremely important for the VQA task. Some work \cite{teney2020unshuffling,agrawal2018don,zellers2018swag} have proved it. For the aesthetic question and answer task based on the whole image, three dimensions of data distribution need to be considered: the distribution of the whole picture and the specific objects in the image in the question, the distribution of pictures with obvious objects, and the overall scenery in the data set, the answer to the question is Distribution under the influence of subjective factors.

For the impact of the first data distribution, we roughly divide the labels of the data set into two categories: "related to specific objects" and "related to the image as a whole", and ensure that the number of images and the number of questions in the two categories are consistent. 

In response to the impact of the second data distribution, we have added two more general and adaptable categories: genre and skill. These two shooting-related categories will be further divided into more than 10 sub-categories, each of which has enough questions and answers. The genre and technique will be able to balance the effects of pictures with obvious objects and overall scenery on the model.

In response to the impact of data distribution in Chapter Four, we add a subjective evaluation. Mark the content of the picture as emotionally related. At the same time, to prevent excessively strong subjective distribution deviations, in the labeling of some attributes, we introduced 7000 additional labeled pictures to constrain the generation of questions and answers.

We show the distributions of the basic aesthetic attribute labels, the genre labels, the subjective labels and photography labels in Figure \ref{fig:distribution}, Figure \ref{fig:genres} and Figure \ref{fig:subjective}.

\subsection{Subjective Labels and Photography Component}

To overcome the possible impact of data distribution, we divide the attributes of images and aesthetics into subjective feelings and objective photography techniques. The subjective description comes from people's direct perception of the image, and the label of this part comes from the pre-training label of CLIP. Including tender, magic, lovely, calm, angry, adventurous, and many other attributes. The reason for selecting these words is that the results of the sample survey show that these words can get a more balanced data distribution.

On the other hand, we provide related attributes of photography, including genres and techniques, which contain multiple subcategories. Same as subjective, we use CLIP to mark images. Some properties are shown in Figure \ref{fig:photo}. Many attributes of this part have the limitation of vague definition. Therefore, each category has a confidence level. Only when the confidence level is high enough, the mark of the picture is considered effective. 

CLIP's image classification model is unsupervised, multiple subcategories can be used as input, and the model is required to classify them. Specifically, some attributes such as "Multiple Exposure" require more professional photography skills, so the confidence threshold for this category is higher; such as "Night Photography", which can better distinguish various attributes, so for this the confidence threshold of the class is low.

\subsection{Bias of Data}

Because of the use of unsupervised models such as CLIP to predict the types of pictures, how to design the distribution of options in answers for pictures is the key. 

Usually, we will give up pictures with too high an accuracy rate and too low an accuracy rate, that is, a picture can be judged as B category on the A attribute, or it is difficult to judge which category it is. We use the design threshold and the accuracy of a certain type of problem model to estimate how the threshold of this type of problem should be, including two parameters, the mean and variance. Only when the mean and the standard deviation of answers' distribution meet the accuracy of this type of baseline model (LXMERT for example) higher than 50 \%, answers contribution is efficacious.

\begin{equation}
	\text{accuracy}(V, Q, A, D(\text{avg}), D(\text{std})) > 0.5
\end{equation}
where $D(\text{avg})$ means the average of the answers' distribution, the $D(\text{std})$ means the standard deviation of the answers' distribution. 72,168 pictures satisfy the above formula.

\begin{table*}[t]
	\centering
	\caption{The \textbf{AoC} in the table means the data with adjustment of image's class confidences, and the \textbf{AoDA} in the table means the adjustment of the distribution of answers for each question. "M1" means method 1, it is the LXMERT, the baseline model. "M2" means method 2, it is the Visual BERT. "M3" means method 3, it is the UNITER, the state-of-the art model in VQA. The training set in the AesVQA database was 58,168 images, and the test set and validation set were both 7,000 images, which were randomly mixed and assigned.}
	\label{tab:freq}
	\resizebox{1.0\textwidth}{!}{
		\begin{tabular}{cccccccccc}
			\hline
			& M1 & M1\&AoC & M1\&AoDA & M2 & M2\&AoC & M2\&AoDA & M3 & M3\&AoC & M3\&AoDA \\ \hline
			Composition      & 53.4\% & 54.6\%      & 54.8\%       & 55.6\%     & 56.3\%          & 56.5\%           & 57.6\% & 59.2\%          & \textbf{60.3\%} \\ \hline
			Color            & 55.3\% & 57.6\%      & 58.8\%       & 57.2\%     & 59.5\%          & 60.2\%           & 58.2\% & 58.5\%          & \textbf{60.5\%} \\ \hline
			Light            & 70.6\% & 73.5\%      & 67.3\%       & 72.5\%     & \textbf{75.8\%} & 69.0\%           & 71.5\% & 72.5\%          & 67.8\%          \\ \hline
			Subject          & 45.9\% & 50.2\%      & 55.5\%       & 48.3\%     & 53.2\%          & 57.1\%           & 49.5\% & 53.9\%          & \textbf{58.0\%} \\ \hline
			Genres           & 48.5\% & 56.6\%      & 54.7\%       & 50.3\%     & 58.5\%          & 56.5\%           & 52.2\% & \textbf{61.2\%} & 60.7\%          \\ \hline
			Techniques       & 42.4\% & 49.9\%      & 50.3\%       & 45.9\%     & 51.2\%          & 53.0\%           & 47.2\% & 53.6\%          & \textbf{54.9\%} \\ \hline
			Subject          & 51.8\% & 52.6\%      & 53.5\%       & 53.9\%     & 54.4\%          & 55.8\%           & 55.6\% & 57.8\%          & \textbf{59.4\%} \\ \hline
			All              & 57.7\% & 58.6\%      & 59.3\%       & 59.8\%     & 60.7\%          & 62.2\%           & 60.3\% & 61.4\%          & \textbf{61.9\%} \\ \hline
	\end{tabular}}
\end{table*}

\begin{table*}[t]
	\centering
	\caption{By randomly extracting 1000 pictures from each category, the confidence adjustment operation needed to be judged. The following table describes the pre-adjusted and post-adjusted range of each sub-category in the categories of genres and techniques:}
	\label{tab:Adjustment}
	\begin{tabular}{ccc}
		\hline
		\textbf{Genres} & Before Adjustment & After Adjustment \\ \hline
		Abstract & 0.536 & 0.536 \\ \hline
		Architectural & 0.912 & 0.615 \\ \hline
		Astrophotography & 0.990 & 0.653 \\ \hline
		Conceptual photography & 0.760 & 0.760 \\ \hline
		Fashion & 0.450 & 0.450 \\ \hline
		Fine-art photography & 0.784 & 0.635 \\ \hline
		High-speed photography & 0.856 & 0.654 \\ \hline
		Landscape & 0.750 & 0.653 \\ \hline
		Nature & 0.803 & 0.664 \\ \hline
		Portrait photography & 0.869 & 0.685 \\ \hline
		Sports & 0.427 & 0.427 \\ \hline
		Still life & 0.752 & 0.645 \\ \hline
		Underwater photography & 0.416 & 0.416 \\ \hline
		Wildlife & 0.947 & 0.703 \\ \hline
		\textbf{Techniques} & Before Adjustment & After Adjustment \\ \hline
		Bokeh & 0.514 & 0.514 \\ \hline
		Burst mode & 0.272 & 0.452 \\ \hline
		Contre-jour & 0.634 & 0.634 \\ \hline
		High-dynamic-range imaging & 0.595 & 0.595 \\ \hline
		Holography & 0.402 & 0.402 \\ \hline
		Long-exposure photography & 0.308 & 0.502 \\ \hline
		Macro photography & 0.869 & 0.652 \\ \hline
		Multiple exposure & 0.894 & 0.675 \\ \hline
		Night photography & 0.613 & 0.613 \\ \hline
		Panning & 0.597 & 0.597 \\ \hline
		Panoramic photography & 0.749 & 0.678 \\ \hline
		Stereoscopy & 0.875 & 0.658 \\ \hline
		Time-lapse photography & 0.605 & 0.605 \\ \hline
		Vignetting & 0.634 & 0.634 \\ \hline
	\end{tabular}
\end{table*}

\begin{figure}[htbp]
	\centering
	\includegraphics[width=0.8\textwidth]{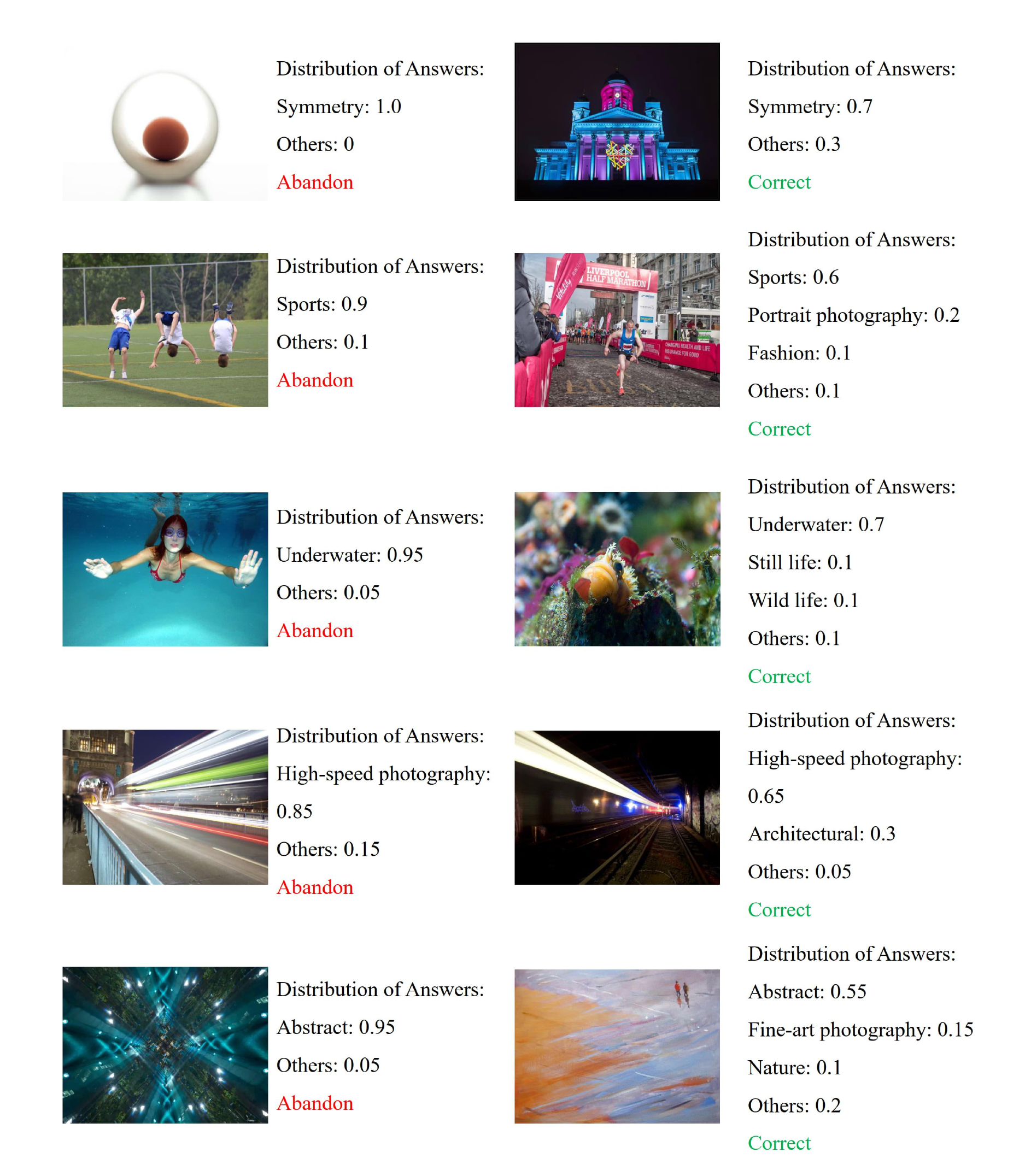}
	\caption{The pictures on the left show that these pictures have more biased attributes, which can easily cause the model to fall into overfitting; the pictures on the right have appropriate "controversial" labels, which can allow the model to obtain better training results.}
	\label{fig:AoDoA}
\end{figure}

\subsection{Influence from Artificial labels}

Complete automatic program marking is not considered to be reliable, and there may also be cases of inconsistency with the facts in automatic program marking. Therefore, we selected 7000 pictures for manual marking. When the result of manual marking does not match the result of machine marking, the picture will be corrected to the result of manual marking.

Another important significance of manually marked pictures is to determine whether the distribution of answers of this type conforms to the real situation. Taking the composition category as an example, the number of symmetrical pictures is usually much lower than that of the rule of thirds and the central composition. This feature can be reflected in the sampled pictures and all pictures. Considering the huge workload of manual marking, we only added manual markings to the composition, color, scene, and lighting attributes. Genres and techniques are more complicated and prone to controversy, so they are not marked.

\section{Experiments}
In the experiment, we used three models to train and test each category. The ratio of the training set and validation set is 8:1, and the number of test sets is the same as the validation set. To test the influence of data distribution on the model, we constantly adjusted the proportions of each sub-category during training and obtained more balanced training results.

The data distribution without artificial interference comes from all pictures. Such a data distribution will not be conducive to the training of the neural network. The operation of adjusting the data set mainly includes: adjusting the confidence threshold of each sub-category in a large category, and the distribution of the question and answer results of the pictures in each large category. The number of pictures is limited by the different types of pictures that have different requirements in the selection process, so it is impossible to achieve close results. The results of the comparative experiment with AoC (Adjustment of Confidence) and AoDA (Adjustment of the Distribution of Answers) are shown in Table 2. It can be seen that UNITER, based on multiple datasets has achieved the best results in AesVQA.

\subsection{Adjustment of Confidence}
Taking genres and techniques as examples, we adjusted the confidence of different categories through training and test results, called it adjustment of confidence (AoC). For different genres and techniques, it is generally believed that the judgment criteria are also different, which is reflected in the labeling process with unsupervised learning as the main method, and the confidence level needs to be adjusted according to the specific class. As is shown in Table 3, The adjusted result is artificially limited to the range of 0.3 to 0.7, this range will help the training and prediction of the model.

\subsection{Adjustment of Distribution of Answers}
Limited to automatic labeling, our proposed dataset requires multiple models to label images. To prevent the occurrence of some over-fitting phenomena, ten answers are set for each question. For adjusting the distribution of answers (AoDA), we let questions have no controversial answers and salient answers. 

The salient answers mean that it is obvious to know which is the right choice in all options. The controversial answers mean there are two or more answers with high confidence. For example, a picture may be classified in "landscape" and "wildlife" at the same time, such a picture needs to be removed, as shown in Figure \ref{fig:photo}.

The adjustment of the distribution of answers will help eliminate images in the data set that lack diversity in labels. These images will affect the model’s preference and make the model more prone to overfitting. The example in the Figure \ref{fig:AoDoA} illustrates which pictures should be discarded and which should be kept during adjustment.

\section{Conclusions}

In this paper, we propose and solve a new task of aesthetic quality evaluation: VQA of image aesthetics. This paper constructs the VQA dataset for image aesthetics. We get basic image aesthetics labels, genres labels, and techniques labels from photography, and subjective emotional labels, and transfer these labels to question-answers pairs. We used three advanced VQA models for training and testing and obtained further performance improvement by adjusting the confidence in image classification and the distribution of answers in VQA. It can be proved by experiments that exporting the answer to aesthetic questions and answers is available by training on our datasets. As a result, display performance model and the advantage of databases, which filled in the blank in the field of aesthetics in the VQA.

\bibliographystyle{ACM-Reference-Format}
\bibliography{sample-base}

\end{document}